\documentclass[10pt,twocolumn,letterpaper]{article}
\usepackage[utf8]{inputenc}
\usepackage[T1]{fontenc}
\usepackage{lmodern}
\usepackage{textcomp}
\DeclareUnicodeCharacter{2212}{\ensuremath{-}}
\DeclareUnicodeCharacter{2192}{\ensuremath{\rightarrow}}
\DeclareUnicodeCharacter{00D7}{\ensuremath{\times}}
\usepackage[margin=0.75in,top=0.75in,bottom=0.85in,columnsep=0.25in]{geometry}
\usepackage[protrusion=true,expansion=false]{microtype}
\usepackage{graphicx}
\usepackage{booktabs}
\usepackage{array}
\usepackage{ragged2e}
\usepackage[font=small,labelfont=bf,skip=4pt]{caption}
\usepackage{enumitem}
\setlist{itemsep=2pt,topsep=3pt,parsep=0pt,leftmargin=1.7em}
\newenvironment{refs}{\par\footnotesize\begin{enumerate}[leftmargin=2.2em,itemsep=2.5pt,topsep=4pt,parsep=0pt]}{\end{enumerate}}
\usepackage{fancyvrb}
\usepackage{balance}
\usepackage{titlesec}
\titleformat{\section}{\normalfont\large\bfseries\raggedright\hyphenpenalty=10000\exhyphenpenalty=10000}{}{0em}{}
\titlespacing*{\section}{0pt}{13pt plus 2pt minus 2pt}{6pt}
\titleformat{\subsection}{\normalfont\normalsize\bfseries\raggedright\hyphenpenalty=10000\exhyphenpenalty=10000}{}{0em}{}
\titlespacing*{\subsection}{0pt}{10pt plus 2pt minus 2pt}{4pt}
\usepackage{xurl}
\usepackage[hidelinks,bookmarksopen=true,bookmarksopenlevel=1,
  pdftitle={Toryx Stratum: Evidence Selection for Local Multi-Document QA},
  pdfauthor={Luis M. Sánchez, Kosrow Dehnad, Marcelo Siero},
  pdfsubject={Technical report},
  pdfkeywords={retrieval-augmented generation, multi-hop question answering, evidence selection, customer-owned hardware, local inference, cost analysis}]{hyperref}
\providecommand{\tightlist}{\setlength{\itemsep}{1.5pt}\setlength{\parskip}{0pt}}
\begin{document}
\setlength{\parindent}{0pt}
\setlength{\parskip}{3.5pt plus 1pt}
\twocolumn[{
\begin{minipage}{\textwidth}
\centering
{\bfseries\fontsize{17}{21}\selectfont Toryx Stratum: Evidence Selection for Local Multi-Document QA\par}
\vspace{6pt}
{\fontsize{10}{12.5}\selectfont\itshape A controlled evaluation of evidence selection, answer accuracy, and input requirements.\par}
\vspace{12pt}
\setlength{\tabcolsep}{0pt}
\begin{tabular}{@{}p{0.33\textwidth}p{0.33\textwidth}p{0.33\textwidth}@{}}
\begin{minipage}[t]{0.94\linewidth}\vspace{0pt}\centering{\normalsize Luis M. Sánchez\textsuperscript{1}}\par\vspace{1.5pt}{\footnotesize\itshape Toryx Inc.}\par\vspace{1pt}{\footnotesize\href{mailto:luis.m.sanchez@toryx.ai}{\ttfamily\small luis.m.sanchez@toryx.ai}}\end{minipage} & \begin{minipage}[t]{0.94\linewidth}\vspace{0pt}\centering{\normalsize Kosrow Dehnad\textsuperscript{2}}\par\vspace{1.5pt}{\footnotesize\itshape Industrial Engineering and Operations Research, Columbia University}\par\vspace{1pt}{\footnotesize\href{mailto:kd11@columbia.edu}{\ttfamily\small kd11@columbia.edu}}\end{minipage} & \begin{minipage}[t]{0.94\linewidth}\vspace{0pt}\centering{\normalsize Marcelo Siero\textsuperscript{3}}\par\vspace{1.5pt}{\footnotesize\itshape Computer Science and Engineering, University of California, Santa Cruz}\par\vspace{1pt}{\footnotesize\href{mailto:msiero2@ucsc.edu}{\ttfamily\small msiero2@ucsc.edu}}\end{minipage}
\end{tabular}\par
\vspace{12pt}
\begin{minipage}{0.94\textwidth}
{\centering\bfseries Abstract\par}
\vspace{5pt}
{\small\RaggedRight\setlength{\parindent}{0pt}\setlength{\parskip}{4pt plus 1pt}
Multi-document question answering requires finding the relevant evidence and then using it to produce an answer. We evaluate Toryx Stratum (STRATUM), an evidence-selection layer coupled to Toryx-Glimmer, an adapted 30-billion-parameter open-weight large language model (LLM). The central question is whether a selected subset of evidence can improve answer accuracy when all required evidence is already present in the larger context.

We test this with a control that keeps the answering model and the questions fixed and changes only the evidence supplied. The control uses development questions that also informed configuration choices, so its results are not held out. Supplying only STRATUM's selected passages (up to ten of roughly 35 to 45 candidates per question) instead of the full candidate set raised exact match (EM) on all three development sets: by 6.0 points on HotpotQA, 3.0 on 2WikiMultiHopQA and 9.0 on MuSiQue. The paired 95\% intervals exclude zero for HotpotQA {[}1.5, 11.0{]} and MuSiQue {[}3.0, 15.0{]} but not for 2WikiMultiHopQA {[}−1.0, 7.5{]}.

On 300 held-out questions per benchmark, STRATUM plus Toryx-Glimmer was competitive with five frontier cloud models on HotpotQA and 2Wiki but trailed on MuSiQue. It scored 69.0\%, 86.3\% and 55.0\% EM, respectively, using roughly one-fifth to one-third as many answering-model input tokens as the cloud models. Coverage diagnostics identify a remaining difficulty with longer evidence chains. Answers cite passage identifiers; an exact answer that cites every gold passage is produced on 66.0\%, 85.7\% and 43.7\% of the held-out questions.\par}
\end{minipage}\vspace{14pt}
\end{minipage}
}]
\pdfbookmark[1]{Abstract}{abstract}
\section*{1. Introduction}
\pdfbookmark[1]{1. Introduction}{1-introduction}

The performance of a document question-answering system depends on more than the language model that writes its answer. A question may name a company while the decisive evidence lies in an ownership filing, a financing agreement and a later amendment. Each passage may be easy to read; finding the passages and connecting them is the system-level problem. A larger context gives the answering model more material to inspect, but does not make that material useful.

This matters for locally deployable models. A system that selects a compact evidence set may reduce the input its answering model needs while keeping the information the task requires. Whether selection preserves or improves accuracy is an empirical question that cannot be inferred from model size or passage count.

Our central question is whether selected evidence can improve accuracy when all required evidence is already available in the larger context. Every candidate set in this study contains the supporting passages (Section 3.1), so the experiments examine both whether selection retains those passages and how accurately the answering model uses the resulting context.

We report two experiments, and they answer different questions.

\begin{itemize}
\tightlist
\item
  \textbf{Development-set control (Section 4).} The answering model, questions and output limit are fixed; only the evidence changes. This isolates the effect of evidence selection within one model. Its questions also informed configuration choices, so its results are not held out.
\item
  \textbf{Held-out system comparison (Section 5).} Paired with Toryx-Glimmer, an adapted open-weight 30B model, STRATUM was competitive with five frontier cloud models on two of three held-out benchmarks. It ranked second on HotpotQA, first on 2Wiki and last on MuSiQue by exact match, using roughly one-fifth to one-third as many answering-model input tokens as the cloud models. The answering model and the evidence condition change together, so this measures system-level performance, not the causal contribution of selection.
\end{itemize}

Two earlier studies motivate this evaluation. \emph{The Commercial Tax} examines the licensing and cost conditions attached to multi-hop retrieval benchmarks and identifies a commercially deployable retrieval foundation, separating the cost of indexing from recurring answering \hyperref[ref:1]{{[}1{]}}. \emph{Clean Scores, Buried Evidence, and Confident Wrong} examines how moving evidence from clean to buried conditions changes agent accuracy, tool use and cost per correct answer \hyperref[ref:2]{{[}2{]}}. Together they motivate evaluating the evidence supplied to a model and the deployment conditions under which results are obtained.

The report provides three connected results: the within-model evidence-selection effect on development sets (Section 4); accuracy and input requirements on three benchmarks in the held-out comparison (Sections 5 and 6); and a rate-based cost calculation tied to recorded token usage (Appendix C).

The practical setting includes private financial, insurance and restricted analytical collections, for example reconciling credit agreements with amendments and disclosures, or comparing policy wording, endorsements and claim documentation. The measurements reported here use controlled public benchmark candidate sets. They are not measured financial or insurance decision outcomes, and the research question is narrower than a production deployment claim.

Retrieval-augmented generation connects language models to external evidence \hyperref[ref:6]{{[}6{]}}, while long-context studies show that supplying more text does not guarantee effective use of it \hyperref[ref:7]{{[}7{]}}. Model adaptation \hyperref[ref:8]{{[}8{]}} and source-cited generation \hyperref[ref:9]{{[}9{]}} address complementary parts of the pipeline. This report evaluates their combination at the evidence-selection and answer interfaces, treating STRATUM as the selection component rather than as an unrestricted corpus-search benchmark.

\section*{2. System overview}
\pdfbookmark[1]{2. System overview}{2-system-overview}

\subsection*{2.1 Evidence selection and answering}
\pdfbookmark[2]{2.1 Evidence selection and answering}{2-1-evidence-selection-and-answering}

Figure 1 separates the deployment workflow from the boundary used in the experiments. In deployment, a document collection supplies candidate passages. In the evaluation, the candidate set is fixed in advance. STRATUM receives a question and the candidate set, returns ranked passages with stable identifiers, and supplies up to ten passages to the answering LLM.

\begin{figure*}[t]
\centering
\includegraphics[width=\textwidth]{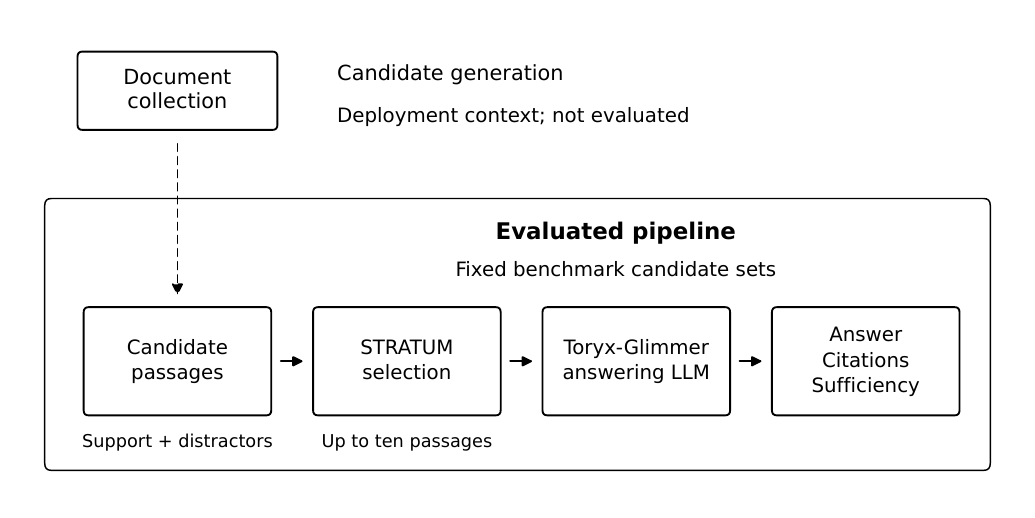}
\caption*{Figure 1. Document-to-answer workflow. The dashed path represents candidate generation from a collection. The evaluated pipeline starts from fixed benchmark candidate sets and uses solid arrows through STRATUM selection, Toryx-Glimmer and the answer output. Candidate sets contain supporting passages and distractors; Section 3 defines their construction. Customer-controlled deployment is the design context, while the accuracy runs used the serving configuration in Section 3.3.}
\end{figure*}

STRATUM\footnote{The first author has filed a U.S. provisional patent application concerning aspects of STRATUM's retrieval architecture (No.~64/010,149, submitted March 19, 2026). This disclosure does not imply a granted patent.} combines retrieval signals in a staged evidence-selection cascade with iterative query refinement. Its commercial configuration uses NVIDIA's Nemotron-3-Embed-8B, following the commercially deployable retrieval foundation examined in \hyperref[ref:1]{{[}1{]}}. The interface exposes ranked passages and their identifiers. Internal ranking weights, prompts and extraction heuristics are not disclosed in this report; the empirical object is the evidence set delivered to the answering model.

Toryx-Glimmer is adapted from Meta's Muse Glimmer 30B. Training uses supervised multi-document examples and STRATUM's evidence-and-answer format across several datasets. A single generalist checkpoint and instruction are used for all three headline benchmarks. The output interface separates the selection component from the answering LLM, although changing either component would require evaluating the resulting configuration.

\subsection*{2.2 Output representation}
\pdfbookmark[2]{2.2 Output representation}{2-2-output-representation}

The model produces numbered steps with passage identifiers, a short answer, a full answer for multi-part questions, an evidence list and a sufficiency judgment. An illustrative benchmark-style output is:

\begin{figure*}[t]
\centering
\begin{minipage}{0.97\textwidth}
\begin{Verbatim}[fontsize=\small]
Question: On what date was the performer of Till dom ensamma born?
Reasoning:
1. The performer is Mauro Scocco, citing [DOC-05].
2. Mauro Scocco was born on 11 September 1962, citing [DOC-21].
Answer: 11 September 1962
Full Answer: 11 September 1962
Evidence: [DOC-05, DOC-21]
Sufficiency: sufficient
\end{Verbatim}
\end{minipage}
\end{figure*}

Stable identifiers let a reviewer inspect the selected source passages. This study scores the final answer and reports evidence-coverage diagnostics; it does not establish that every cited passage entails its associated step. The sufficiency field lets the model decline to answer, but this report does not establish that its sufficiency judgment is reliable. An \texttt{INSUFFICIENT} response is an abstention and is scored as incorrect. Section 5.3 also reports answer coverage and exact match conditional on answering, without changing the benchmark denominator.

\section*{3. Experimental design}
\pdfbookmark[1]{3. Experimental design}{3-experimental-design}

\subsection*{3.1 Questions and candidate sets}
\pdfbookmark[2]{3.1 Questions and candidate sets}{3-1-questions-and-candidate-sets}

We use three public multi-document question-answering benchmarks. HotpotQA \hyperref[ref:3]{{[}3{]}} asks questions requiring evidence from multiple Wikipedia articles, including linking and comparison questions. 2WikiMultiHopQA \hyperref[ref:4]{{[}4{]}} constructs questions with explicit multi-hop evidence paths across Wikipedia sources. MuSiQue \hyperref[ref:5]{{[}5{]}} composes multi-hop questions from connected single-hop questions and is designed to reduce shortcuts that bypass the required reasoning chain. A hop is a dependency between facts needed to answer a question, not a count of model calls. These benchmarks test evidence combination rather than an operational finance, insurance or defense workflow.

Each benchmark has a fixed 200-question development set and a separate 300-question locked test. Development sets select the configuration; the locked-test questions are excluded from training and configuration selection, and each locked test is run once per evaluated configuration. HotpotQA and 2Wiki questions were sampled from a 1,000-question pool per benchmark, stratified by question type without score-based selection. The MuSiQue questions come from the 1,000-question MuSiQue subset released with HippoRAG 2 {[}12{]}. Its locked test is 300 questions drawn by seeded simple random sampling from the first 500 questions of that file, without score-based selection; the list was frozen before the candidate sets were built. The other 700 questions were divided by a seeded draw, stratified by hop count, into the 200-question development set and 500 questions reserved for training. The locked test contains 163 two-hop, 89 three-hop and 48 four-hop questions; the development set contains 101, 65 and 34.

Each question has a frozen candidate set: its benchmark-supplied supporting passages and distractors, supplemented with nearest-neighbour passages from the benchmark paragraph library. Mean sizes are 34.94 passages for HotpotQA, 34.96 for 2Wiki and 44.99 for MuSiQue. Every candidate set includes the required supporting passages. STRATUM selects up to ten; the headline cloud comparators receive all candidates. The experiment therefore measures selection and answering within the same supplied candidate sets, not retrieval over an unrestricted collection. Open-index studies such as UNREAL \hyperref[ref:10]{{[}10{]}} use a different evaluation boundary.

The generalist and STRATUM components use public training-split data. The benchmark families are not unseen by the pipeline, and their public text may occur in upstream model pretraining. Held-out exclusion refers to the training and selection performed for this study.

\subsection*{3.2 Comparators and evidence conditions}
\pdfbookmark[2]{3.2 Comparators and evidence conditions}{3-2-comparators-and-evidence-conditions}

The frontier panel comprises Gemini 3.8 Flash, Muse Spark 1.3, Grok 4.6, GPT-6.1 and Claude Opus 5.5. These were the five highest-scoring models by MuSiQue locked-test EM among eleven tested on October 5, 2026, before the HotpotQA and 2Wiki results existed; the same five led by F1. The other six scored 31.3--49.7 EM on MuSiQue and were not carried forward. This selection defines the comparison panel; it is not an exhaustive ranking of available models.

The within-model control in Section 4 varies only the evidence: gold passages, STRATUM's top eight, its up-to-ten selection, or all candidates. Gold-only is an oracle condition, not a retrieval output.

\subsection*{3.3 Serving and evaluation metrics}
\pdfbookmark[2]{3.3 Serving and evaluation metrics}{3-3-serving-and-evaluation-metrics}

The open-weight model was trained and served on an NVIDIA A100-80GB. Serving used vLLM, bf16 and greedy decoding, with maximum outputs of 8,192 tokens for locked tests and 2,048 for development runs. Cloud models used the provider routes in Appendix A; temperature was zero where supported, otherwise the provider default. Reasoning settings followed the available routes, including reasoning-enabled GPT-6.1 and Opus calls where no off mode was available.

Exact match (EM) compares the normalized final answer with the reference and accepted aliases. HotpotQA and 2Wiki use their official answer scorers; MuSiQue uses SQuAD-style normalization. Token F1 measures overlap with the reference and credits partial answers. Input tokens are the mean prompt tokens reported by each server or API; tokenizers differ across models.

API errors, empty replies, truncated replies and \texttt{INSUFFICIENT} outputs score zero and remain in the denominator. Failed requests were retried with back-off; remaining failures count as incorrect. Paired differences use 10,000 percentile bootstrap resamples over questions, with exact two-sided McNemar tests for EM-discordant pairs. Intervals are marginal, without multiplicity correction. A nonsignificant difference is not evidence of equivalence.

\section*{4. Evidence-selection effect (development control)}
\pdfbookmark[1]{4. Evidence-selection effect (development control)}{4-evidence-selection-effect-development-control}

We evaluate the same generalist on each benchmark's 200 development questions with four inputs: only gold passages, STRATUM's top eight, its up-to-ten selection, and all candidates. The answering model, question set and output cap are fixed.

\begin{figure*}[t]
\centering
\includegraphics[width=\textwidth]{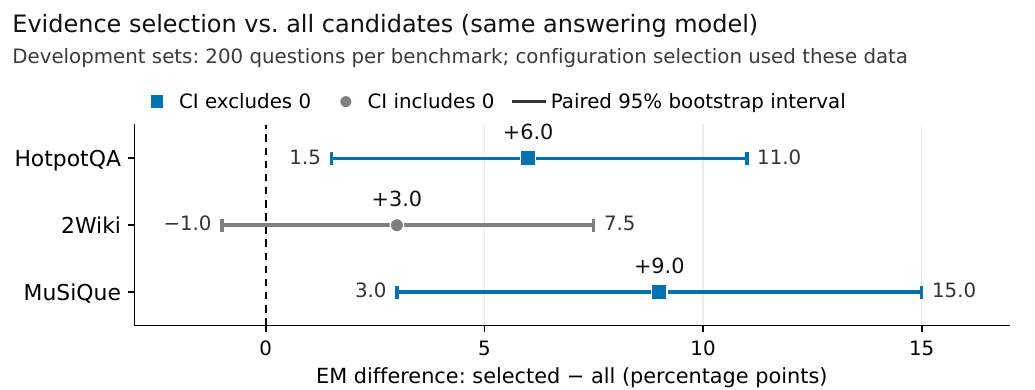}
\caption*{Figure 2. Same-model evidence-selection effect on 200 development questions per benchmark: EM with STRATUM-selected passages minus EM with all candidates. Markers show paired estimates and bars show 95\% bootstrap intervals. The intervals exclude zero on HotpotQA and MuSiQue, but not on 2Wiki. These data informed configuration selection.}
\end{figure*}

\begin{table*}[t]
\centering
\small
\setlength{\tabcolsep}{4.0pt}
\renewcommand{\arraystretch}{1.14}
\captionof*{table}{\textbf{Table 1. Within-model evidence-budget results, development-set EM (\%).} Gold-only is an oracle condition. All other columns use the same generalist and questions.}
\vspace{1pt}
\begin{tabular}{@{}l r r r r@{}}
\toprule
Benchmark & Gold only & STRATUM top eight & STRATUM up to ten & All candidates \\
\midrule
HotpotQA & 75.5 & 70.5 & 70.0 & 64.0 \\
2Wiki & 84.0 & 82.0 & 81.0 & 78.0 \\
MuSiQue & 68.0 & 56.5 & 61.5 & 52.5 \\
\bottomrule
\end{tabular}
\end{table*}

Replacing all candidates with STRATUM's selection raises EM by 6.0 points on HotpotQA (95\% CI {[}1.5, 11.0{]}), 3.0 on 2Wiki {[}−1.0, 7.5{]} and 9.0 on MuSiQue {[}3.0, 15.0{]}. The paired evidence supports an improvement on two of the three development sets. The full candidate set includes every gold passage, yet it produces lower observed scores than the selection on all three tasks. Having the evidence available and using it effectively are different properties of the system.

Increasing the selection from eight passages to ten raises MuSiQue EM from 56.5 to 61.5, a paired difference of +5.0 points (95\% CI {[}1.0, 9.5{]}; exact McNemar p = 0.041). The corresponding differences on HotpotQA and 2Wiki are −0.5 {[}−3.5, 2.5{]} and −1.0 {[}−3.0, 1.0{]}; both intervals include zero. Selection therefore involves a budget trade-off: excluding distractors can help, while too small a budget can exclude a required link.

\section*{5. Held-out system comparison}
\pdfbookmark[1]{5. Held-out system comparison}{5-held-out-system-comparison}

\subsection*{5.1 Answer accuracy}
\pdfbookmark[2]{5.1 Answer accuracy}{5-1-answer-accuracy}

Table 2 reports the held-out comparison. The combined STRATUM--Toryx-Glimmer system achieves 69.0 EM on HotpotQA, 86.3 on 2Wiki and 55.0 on MuSiQue, ranking second, first and last among the six systems, respectively. Figure 3 shows the individual locked-test scores on a common 0--100 scale, without aggregation.

\begin{table*}[t]
\centering
\small
\setlength{\tabcolsep}{4.0pt}
\renewcommand{\arraystretch}{1.14}
\captionof*{table}{\textbf{Table 2. Locked-test exact match, with token F1 in parentheses (\%).} Each column contains the same 300 questions for all systems. STRATUM supplies up to ten passages to Toryx-Glimmer; cloud models receive all candidates.}
\vspace{1pt}
\begin{tabular}{@{}l r r r@{}}
\toprule
System & HotpotQA & 2Wiki & MuSiQue \\
\midrule
STRATUM + Toryx-Glimmer & 69.0 (82.0) & 86.3 (89.7) & 55.0 (61.8) \\
Gemini 3.8 Flash & 68.7 (82.6) & 82.7 (90.0) & 67.0 (77.4) \\
Muse Spark 1.3 & 69.7 (81.9) & 83.3 (90.6) & 66.7 (78.6) \\
Grok 4.6 & 64.0 (79.1) & 82.0 (90.8) & 62.3 (78.0) \\
GPT-6.1 & 62.3 (78.9) & 74.0 (81.5) & 59.3 (75.5) \\
Claude Opus 5.5 & 63.3 (80.1) & 78.3 (85.0) & 58.0 (74.5) \\
\bottomrule
\end{tabular}
\end{table*}

\begin{figure*}[t]
\centering
\includegraphics[width=\textwidth]{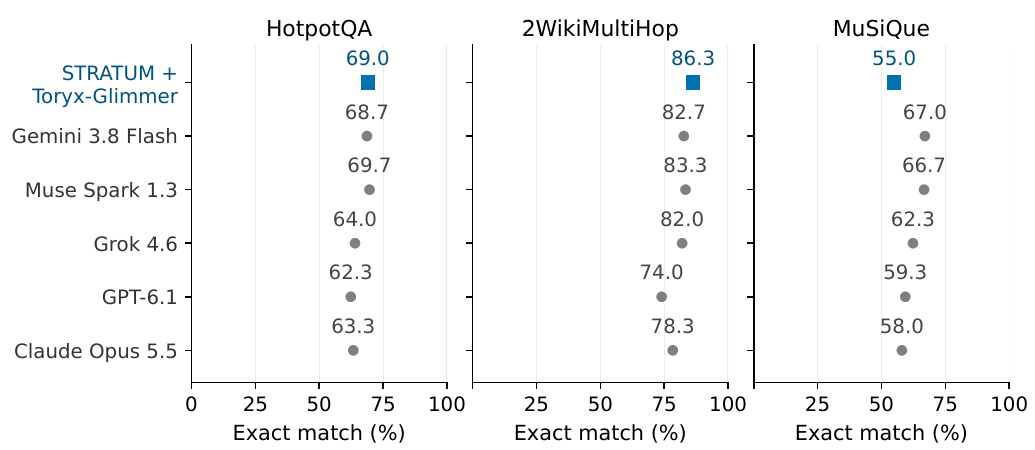}
\caption*{Figure 3. Individual-system EM on 300 held-out questions per benchmark, using a common 0--100 scale and system order. The blue square denotes STRATUM with Toryx-Glimmer; gray circles denote each cloud model. Evidence conditions follow Section 3.2. Paired comparisons are in Appendix A.}
\end{figure*}

On HotpotQA, the system is not statistically distinguishable from Gemini or Muse, the two leading cloud configurations in this sample. Its EM advantage over GPT-6.1 is supported by the paired comparison.

On 2Wiki, it exceeds GPT-6.1 by 12.3 EM points and Opus by 8.0 points; both paired intervals exclude zero. Its differences from Gemini and Muse are unresolved. Its F1 is not distinguishable from the three strongest cloud models, which indicates that part of the EM advantage concerns exact answer form.

On MuSiQue, the system trails Gemini, Muse and Grok with paired EM intervals excluding zero. The EM differences from GPT-6.1 and Opus are unresolved, while F1 is below all five cloud configurations. The five panel models are the top five of eleven on this benchmark (Section 3.2); the six not carried forward scored 31.3--49.7 EM, below the system's 55.0.

The full paired statistics are in Appendix A. The system is competitive on two tasks and trails on the longer-chain benchmark; the results do not give a uniform ranking.

\subsection*{5.2 Answering-model input requirements}
\pdfbookmark[2]{5.2 Answering-model input requirements}{5-2-answering-model-input-requirements}

Table 3 records the input associated with those results. The generalist receives approximately 1,500--1,850 prompt tokens per question, compared with approximately 5,100--9,700 for the cloud panel. The reduction is approximately three- to five-fold on HotpotQA and 2Wiki, where the system is in the leading group; across all comparisons its input is roughly one-fifth to one-third of the cloud range. This is an input-workload result, not an end-to-end speed or energy measurement.

\begin{table*}[t]
\centering
\small
\setlength{\tabcolsep}{4.0pt}
\renewcommand{\arraystretch}{1.14}
\captionof*{table}{\textbf{Table 3. Mean answering-LLM input tokens per question, locked tests.} Counts cover the answer request, not other retrieval stages, and are tokenizer-dependent.}
\vspace{1pt}
\begin{tabular}{@{}l r r r@{}}
\toprule
System & HotpotQA & 2Wiki & MuSiQue \\
\midrule
STRATUM + Toryx-Glimmer & 1,537 & 1,838 & 1,676 \\
Gemini 3.8 Flash & 5,357 & 5,705 & 6,425 \\
Muse Spark 1.3 & 5,251 & 5,567 & 6,264 \\
Grok 4.6 & 5,110 & 5,408 & 6,066 \\
GPT-6.1 & 5,496 & 5,514 & 6,690 \\
Claude Opus 5.5 & 8,089 & 8,456 & 9,678 \\
\bottomrule
\end{tabular}
\end{table*}

\subsection*{5.3 Evidence completeness and longer chains}
\pdfbookmark[2]{5.3 Evidence completeness and longer chains}{5-3-evidence-completeness-and-longer-chains}

The selected set contains every gold supporting passage for 292 of 300 HotpotQA questions (97.3\%), 296 of 300 2Wiki questions (98.7\%) and 245 of 300 MuSiQue questions (81.7\%). On MuSiQue, EM is 64.5\% when the selection is complete and 12.7\% when it is not. The system also returns \texttt{INSUFFICIENT} on 39 MuSiQue questions, compared with five and three on the other tasks; all are scored incorrect.

\begin{table*}[t]
\centering
\small
\setlength{\tabcolsep}{4.0pt}
\renewcommand{\arraystretch}{1.14}
\captionof*{table}{\textbf{Table 4. Coverage and output diagnostics for the generalist, locked tests.} Grounded EM requires an exact answer and citation of every gold passage; it does not verify citation entailment.}
\vspace{1pt}
\begin{tabular}{@{}l r r r r@{}}
\toprule
Benchmark & Complete selection & EM: complete / incomplete & Insufficient answers & Grounded EM \\
\midrule
HotpotQA & 97.3\% (292/300) & 70.9 / 0.0 & 5 & 66.0 \\
2Wiki & 98.7\% (296/300) & 87.5 / 0.0 & 3 & 85.7 \\
MuSiQue & 81.7\% (245/300) & 64.5 / 12.7 & 39 & 43.7 \\
\bottomrule
\end{tabular}
\end{table*}

Coverage is one important part of the longer-chain problem. This is an association, not a causal decomposition: questions with complete and incomplete selections can differ in difficulty, and the complete subset is not comparable to another model's score over the whole test. Both evidence coverage and the model's use of evidence contribute to the system result.

\begin{table*}[t]
\centering
\small
\setlength{\tabcolsep}{4.0pt}
\renewcommand{\arraystretch}{1.14}
\captionof*{table}{\textbf{Table 5. Abstention diagnostics for the generalist, locked tests.} Coverage is the fraction of all 300 questions with a non-\texttt{INSUFFICIENT} answer. Conditional EM is exact match among those answers.}
\vspace{1pt}
\begin{tabular}{@{}l r r r r@{}}
\toprule
Benchmark & Answered / total & Coverage (\%) & EM among answered (\%) & All-question EM (\%) \\
\midrule
HotpotQA & 295/300 & 98.3 & 70.2 & 69.0 \\
2Wiki & 297/300 & 99.0 & 87.2 & 86.3 \\
MuSiQue & 261/300 & 87.0 & 63.2 & 55.0 \\
\bottomrule
\end{tabular}
\end{table*}

These counts were recomputed from the retained per-question outputs. The three generalist runs had no API errors, empty responses or truncated responses. Conditional EM is descriptive: it does not establish calibrated abstention, and it does not replace the all-question scores when comparing systems.

\section*{6. Accuracy and answering-model input}
\pdfbookmark[1]{6. Accuracy and answering-model input}{6-accuracy-and-answering-model-input}

Figure 4 plots the locked-test EM of Table 2 against the mean prompt tokens of Table 3. Each panel is one benchmark; each marker is one of the six systems on the same 300 held-out questions.

\begin{figure*}[t]
\centering
\includegraphics[width=\textwidth]{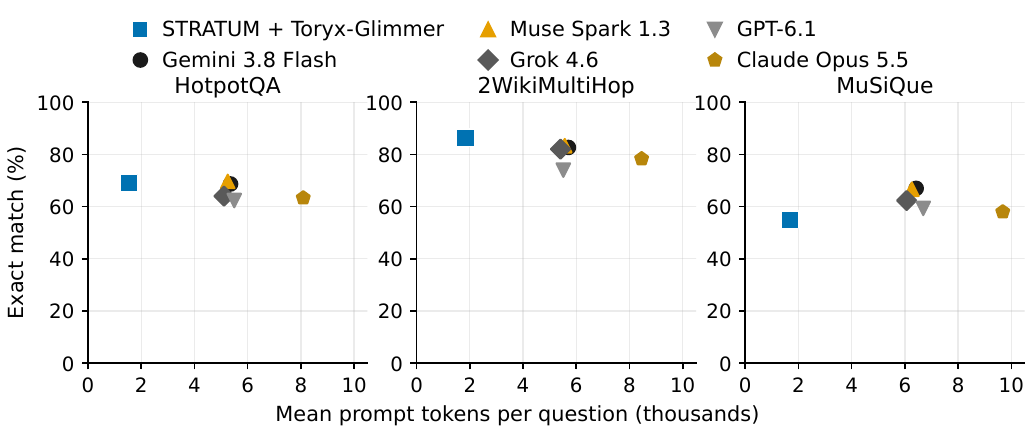}
\caption*{Figure 4. Answer accuracy versus mean answering-model input on the held-out questions, from Tables 2 and 3. The blue square denotes STRATUM with Toryx-Glimmer; other markers identify individual cloud configurations. Tokens are tokenizer-dependent answer-request inputs, not end-to-end compute or latency measurements.}
\end{figure*}

The combined system uses approximately 1,500--1,850 prompt tokens per question while placing in the leading group on HotpotQA and 2Wiki. On MuSiQue it processes similarly small inputs but trails three cloud models, so input reduction alone does not determine accuracy. The cloud points use all candidates; this is not an optimization study of cloud context budgets.

Rate-based cost projections that apply recorded usage to list prices are in Appendix C. They are projections, not measured local operating costs.

\section*{7. Discussion and deployment implications}
\pdfbookmark[1]{7. Discussion and deployment implications}{7-discussion-and-deployment-implications}

\subsection*{7.1 What the two comparisons establish}
\pdfbookmark[2]{7.1 What the two comparisons establish}{7-1-what-the-two-comparisons-establish}

The held-out experiment shows that the combined selection-and-answering system can reach the leading group on two multi-document tasks while supplying substantially less input to its answering LLM. The outcome is task-dependent: the same configuration trails on MuSiQue. Because that experiment changes the model and the evidence together, it does not show what selection contributes. The within-model control does: it provides separate evidence that selection itself can improve accuracy relative to supplying all candidate passages, on development data that were not held out. Neither experiment substitutes for the other.

The evidence-budget result is the principal mechanism-level finding. A context containing every supporting passage is not necessarily the context from which this LLM answers most accurately. Conversely, the MuSiQue coverage result shows that selection must preserve the chain, not merely shorten the prompt. An input-efficient system has to do both.

\subsection*{7.2 Customer-controlled hardware}
\pdfbookmark[2]{7.2 Customer-controlled hardware}{7-2-customer-controlled-hardware}

We target customer-controlled deployments in which model adaptation uses NVIDIA platforms and inference can use Apple silicon or other locally administered hardware. Apple Mac Studio and MacBook Pro configurations, DGX Spark systems and an RTX A6000 workstation are design targets listed in Appendix B. The accuracy measurements in this report were obtained on an A100-80GB, not on those targets.

This separation lets model development and routine inference use different hardware. It is relevant to organizations such as family offices that administer private document collections, and to restricted facilities where document and query processing must remain inside an authorized environment. Where security policy requires air-gapped operation, external model APIs are not a deployable alternative: computation must remain on approved hardware inside the isolated environment. Where external APIs are permitted, the choice between cloud and local execution is a cost and policy comparison. Memory, precision, concurrency and the allocation of retrieval stages determine the practical configuration. Hardware-specific latency, throughput, energy and quantized accuracy require measurement on that configuration.

Defense deployments require an answering model whose applicable terms permit the intended mission, together with authorization for the deployment environment. The answering-model interface permits a substitute model, but the accuracy reported here pertains only to the evaluated Toryx-Glimmer configuration. The benchmark results are research measurements, not certification of a fielded defense configuration.

\subsection*{7.3 Scope of inference}
\pdfbookmark[2]{7.3 Scope of inference}{7-3-scope-of-inference}

The fixed-candidate protocol isolates evidence selection and answering with reference evidence available. It does not measure candidate discovery over a full collection. Public Wikipedia-based tasks with short answers also differ from questions over mixed financial, insurance or legal documents. The generalist's data exposure, comparator selection and development-set controls should be read under the protocol in Section 3.

Exact match and gold-passage citation counts do not establish step-level entailment, and input-token reductions do not measure end-to-end compute, energy or latency. The vendor conducted the evaluation, and independent reproduction has not been established.

\subsection*{7.4 Governance implications}
\pdfbookmark[2]{7.4 Governance implications}{7-4-governance-implications}

Passage identifiers support traceability by letting a reviewer locate the cited evidence. The sufficiency field and explicit abstention option make the model's stated evidence assessment visible; this report does not validate its calibration or demonstrate a reduction in harmful decisions.

Customer-controlled deployment can keep documents, queries, model execution and retained run records within an organization-administered environment. This is a deployment objective, not a measured security property of the benchmark. Whole-system disconnected operation, access control and retention policy require validation in the intended configuration.

These properties can support review of model-assisted financial analysis, documentation of AI-supported insurance decisions, and traceable staff work in defense. They support human inspection; they do not by themselves establish regulatory compliance, operational suitability or the correctness of a decision. A source-linked brief informs, rather than replaces, the responsible analyst or officer.

\section*{8. Conclusion}
\pdfbookmark[1]{8. Conclusion}{8-conclusion}

STRATUM with an adapted 30B LLM achieves competitive held-out accuracy on HotpotQA and 2Wiki with substantially smaller answering inputs, while its MuSiQue result reveals a remaining longer-chain gap. A same-LLM development control shows that STRATUM-selected evidence scored higher than the full candidate set on all three benchmarks, with paired intervals excluding zero on two, separating the value of evidence selection from differences between models.

The results support evaluating the evidence budget as a system-design variable alongside model capability. For locally deployable question answering, the relevant question is not only which model to run, but which passages to put in front of it. Explicit workload and cost accounting connects that question to deployment while keeping target hardware, projected API charges and measured system performance distinct.

\section*{Disclosure of AI assistance}
\pdfbookmark[1]{Disclosure of AI assistance}{disclosure-of-ai-assistance}

AI assistants were used to support software development, experimental analysis, manuscript drafting and revision, and figure preparation. Manuscript review and revision included assistance from Anthropic's Claude and OpenAI's GPT-family assistants. AI-generated suggestions were treated as provisional and checked against source documents, code or retained experimental artifacts as appropriate. The reported empirical results derive from the recorded benchmark runs and scoring procedures, not from an assistant's assessment. The authors take responsibility for the study design, interpretation and final manuscript. This disclosure does not imply independent human verification of every statement; the reproducibility and verification boundaries are described in Section 7.3 and Appendix C.4.

\section*{References}
\pdfbookmark[1]{References}{references}

\begin{refs}
\item \phantomsection\label{ref:1} Sánchez, Luis M., and Kosrow Dehnad. (2026). \emph{The Commercial Tax: Rent-vs-Own Blind Spots in Multi-Hop Retrieval Benchmarks.} arXiv:2608.16096. \url{https://doi.org/10.48550/arXiv.2608.16096}. Replication artifacts: \url{https://doi.org/10.5281/zenodo.21972866}.
\item \phantomsection\label{ref:2} Sánchez, Luis M. (2026). \emph{Clean Scores, Buried Evidence, and Confident Wrong: A Receipt-Based Audit of Frontier Agentic QA.} arXiv:2609.15319. \url{https://doi.org/10.48550/arXiv.2609.15319}. Evidence archive: \url{https://doi.org/10.5281/zenodo.22310532}.
\item \phantomsection\label{ref:3} Yang, Z., Qi, P., Zhang, S., Bengio, Y., Cohen, W. W., Salakhutdinov, R., and Manning, C. D. (2018). \emph{HotpotQA: A Dataset for Diverse, Explainable Multi-hop Question Answering.} EMNLP 2018. arXiv:1809.09600.
\item \phantomsection\label{ref:4} Ho, X., Duong Nguyen, A.-K., Sugawara, S., and Aizawa, A. (2020). \emph{Constructing A Multi-hop QA Dataset for Comprehensive Evaluation of Reasoning Steps.} COLING 2020. arXiv:2011.01060.
\item \phantomsection\label{ref:5} Trivedi, H., Balasubramanian, N., Khot, T., and Sabharwal, A. (2022). \emph{MuSiQue: Multihop Questions via Single-hop Question Composition.} TACL 10, 539--554.
\item \phantomsection\label{ref:6} Lewis, P., Perez, E., Piktus, A., et al.~(2020). \emph{Retrieval-Augmented Generation for Knowledge-Intensive NLP Tasks.} NeurIPS 2020. arXiv:2005.11401.
\item \phantomsection\label{ref:7} Liu, N. F., Lin, K., Hewitt, J., et al.~(2024). \emph{Lost in the Middle: How Language Models Use Long Contexts.} TACL 12, 157--173.
\item \phantomsection\label{ref:8} Hu, E. J., Shen, Y., Wallis, P., et al.~(2022). \emph{LoRA: Low-Rank Adaptation of Large Language Models.} ICLR 2022. arXiv:2106.09685.
\item \phantomsection\label{ref:9} Gao, T., Yen, H., Yu, J., and Chen, D. (2023). \emph{Enabling Large Language Models to Generate Text with Citations.} EMNLP 2023. arXiv:2305.14627.
\item \phantomsection\label{ref:10} Kinderman, E., Hoffer, E., Blau, Y., Chmiel, B., Banner, R., Soudry, D., and Ginsburg, B. (2026). \emph{UNREAL: Unifying Retrieval and Long-Context with a Single Model.} arXiv:2610.08463.
\item \phantomsection\label{ref:11} Apple. (2026). \emph{Mac Studio, M5 Ultra, 30-core CPU, 64-core GPU, 256 GB unified memory, 1 TB storage.} U.S. online-store list price, accessed November 18, 2026. \url{https://www.apple.com/shop/buy-mac/mac-studio/m5-ultra-chip-30-core-cpu-64-core-gpu-256gb-memory-1tb-storage}
\item \phantomsection\label{ref:12} Gutiérrez, B. J., Shu, Y., Qi, W., Zhou, S., and Su, Y. (2025). \emph{From RAG to Memory: Non-Parametric Continual Learning for Large Language Models.} ICML 2025. arXiv:2502.14802.
\end{refs}

\section*{Appendix A. Comparator configuration and full paired statistics}
\pdfbookmark[1]{Appendix A. Comparator configuration and full paired statistics}{appendix-a-comparator-configuration-and-full-paired-statistics}

\subsection*{A.1 Access routes and scoring details}
\pdfbookmark[2]{A.1 Access routes and scoring details}{a-1-access-routes-and-scoring-details}

\begin{table*}[t]
\centering
\small
\setlength{\tabcolsep}{4.0pt}
\renewcommand{\arraystretch}{1.14}
\begin{tabular}{@{}l l l l@{}}
\toprule
Comparator & Provider route & Model identifier & Reasoning setting \\
\midrule
Gemini 3.8 Flash & Google Vertex AI & \texttt{gemini-3.8-flash} & Thinking budget 0 \\
GPT-6.1 & Amazon Bedrock, us-east-1 & \texttt{us.openai.gpt-6.1-sol} & On; no off mode on the route used \\
Grok 4.6 & Azure AI Foundry & \texttt{grok-4-6} & Provider-supported configuration \\
Muse Spark 1.3 & Azure AI Foundry & \texttt{muse-spark-1.3} & Provider-supported configuration \\
Claude Opus 5.5 & Amazon Bedrock, us-east-1 & \texttt{us.anthropic.claude-opus-5-5} & On; no off mode on the route used \\
\bottomrule
\end{tabular}
\end{table*}

Frontier MuSiQue runs date to October 5; HotpotQA/2Wiki runs to October 14; generalist development and cost runs to October 17--18; and generalist locked tests to October 22, 2026. Scoring uses \texttt{hotpot\_\allowbreak{}evaluate\_\allowbreak{}v1}, \texttt{2wikimultihop\_\allowbreak{}evaluate\_\allowbreak{}v1.1} with its alias file, and SQuAD-style normalization for MuSiQue. The final answer is taken from the last \texttt{Answer:} line after removing reasoning blocks; cited document IDs are parsed from the last \texttt{Evidence:} line.

\subsection*{A.2 Paired comparisons}
\pdfbookmark[2]{A.2 Paired comparisons}{a-2-paired-comparisons}

\begin{table*}[t]
\centering
\footnotesize
\setlength{\tabcolsep}{4.0pt}
\renewcommand{\arraystretch}{1.14}
\captionof*{table}{\textbf{Table A1. STRATUM + Toryx-Glimmer minus each cloud comparator.} Differences are percentage points with 95\% paired bootstrap intervals. McNemar p-values are exact and two-sided; no multiplicity correction is applied.}
\vspace{1pt}
\begin{tabular}{@{}l l l l r@{}}
\toprule
Benchmark & Comparator & EM difference {[}95\% CI{]} & F1 difference {[}95\% CI{]} & McNemar p \\
\midrule
HotpotQA & Gemini 3.8 Flash & +0.3 {[}−5.0, +5.7{]} & −0.6 {[}−4.1, +3.0{]} & 1.000 \\
HotpotQA & Muse Spark 1.3 & −0.7 {[}−5.7, +4.7{]} & +0.1 {[}−3.6, +3.9{]} & 0.899 \\
HotpotQA & Grok 4.6 & +5.0 {[}−0.3, +10.7{]} & +3.0 {[}−0.8, +6.9{]} & 0.096 \\
HotpotQA & GPT-6.1 & +6.7 {[}+0.7, +12.7{]} & +3.1 {[}−0.9, +7.3{]} & 0.040 \\
HotpotQA & Claude Opus 5.5 & +5.7 {[}+0.0, +11.3{]} & +2.0 {[}−1.8, +5.8{]} & 0.064 \\
2Wiki & Gemini 3.8 Flash & +3.7 {[}−0.3, +7.7{]} & −0.3 {[}−3.4, +2.8{]} & 0.117 \\
2Wiki & Muse Spark 1.3 & +3.0 {[}−1.3, +7.3{]} & −0.9 {[}−4.1, +2.2{]} & 0.222 \\
2Wiki & Grok 4.6 & +4.3 {[}+0.0, +9.0{]} & −1.2 {[}−4.2, +1.9{]} & 0.079 \\
2Wiki & GPT-6.1 & +12.3 {[}+7.0, +17.7{]} & +8.2 {[}+3.8, +12.6{]} & \textless0.001 \\
2Wiki & Claude Opus 5.5 & +8.0 {[}+3.0, +13.0{]} & +4.7 {[}+0.7, +8.7{]} & 0.002 \\
MuSiQue & Gemini 3.8 Flash & −12.0 {[}−18.0, −6.0{]} & −15.6 {[}−20.9, −10.4{]} & \textless0.001 \\
MuSiQue & Muse Spark 1.3 & −11.7 {[}−18.0, −5.3{]} & −16.8 {[}−22.0, −11.5{]} & \textless0.001 \\
MuSiQue & Grok 4.6 & −7.3 {[}−13.7, −1.3{]} & −16.2 {[}−21.9, −10.6{]} & 0.026 \\
MuSiQue & GPT-6.1 & −4.3 {[}−10.7, +2.0{]} & −13.7 {[}−19.2, −8.3{]} & 0.208 \\
MuSiQue & Claude Opus 5.5 & −3.0 {[}−9.0, +3.0{]} & −12.7 {[}−17.9, −7.4{]} & 0.386 \\
\bottomrule
\end{tabular}
\end{table*}

The generalist had no API errors or truncated outputs on the three locked tests. Mean output tokens were 130 on HotpotQA, 165 on 2Wiki and 154 on MuSiQue. Full per-configuration records include EM, F1, grounded EM, errors and token counts.

\section*{Appendix B. Deployment design targets}
\pdfbookmark[1]{Appendix B. Deployment design targets}{appendix-b-deployment-design-targets}

The intended hardware configurations separate adaptation and routine inference. They describe customer-controlled deployment targets, not additional benchmark platforms.

\noindent\begin{minipage}{\columnwidth}\centering
\footnotesize
\setlength{\tabcolsep}{4.0pt}
\renewcommand{\arraystretch}{1.14}
\begin{tabular}{@{}>{\raggedright\arraybackslash}p{52.8pt} >{\raggedright\arraybackslash}p{78.4pt} >{\raggedright\arraybackslash}p{81.2pt}@{}}
\toprule
Platform & Configuration & Intended role \\
\midrule
Apple Mac Studio M5 & One system, 256 GB unified memory & Primary Apple inference target \\
Apple MacBook Pro M3 & 36 GB unified memory & Portable development and quantized inference \\
NVIDIA DGX Spark & One or two systems, 128 GB unified memory each & Model development, training and NVIDIA inference \\
NVIDIA RTX A6000 workstation & 48 GB GPU memory; 128 GB system RAM & CUDA development and quantized inference \\
\bottomrule
\end{tabular}
\end{minipage}\par\vspace{7pt plus 1pt}

The complete inventory is not a minimum system requirement. Model precision, context size, simultaneous requests and other resident services determine memory fit. GPU memory, system RAM and unified memory are separate capacities. Measured hardware-specific serving performance is outside this report's accuracy and input-budget evaluation.

\section*{Appendix C. Supplementary cost accounting and evidence record}
\pdfbookmark[1]{Appendix C. Supplementary cost accounting and evidence record}{appendix-c-supplementary-cost-accounting-and-evidence-record}

\subsection*{C.1 Workload and cost boundaries}
\pdfbookmark[2]{C.1 Workload and cost boundaries}{c-1-workload-and-cost-boundaries}

The workload is one million questions per year for three years; the benchmark rows represent alternative question mixes, not additive volumes. For the cloud side, we apply the recorded Opus 5.5 rates of \$4 per million input tokens and \$20 per million output tokens to observed mean answer-request usage. For Q questions, mean input P and mean output C, the projected charge is:

\[
C_{\mathrm{API}}(Q)=\frac{Q}{10^6}\left(4P+20C\right).
\]

The local hardware illustration uses Apple's U.S. list price of \$9,499 for a Mac Studio with M5 Ultra, a 30-core CPU, a 64-core GPU, 256 GB unified memory and 1 TB storage, observed on November 18, 2026 {[}11{]}. A straight-line allocation over three years is \$3,166.33 per year. This is a purchase-price allocation only: it excludes tax, financing, residual value, electricity, software, storage expansion, maintenance, support and training compute. It is not an invoice, an operating-cost measurement or a claim that this machine has demonstrated the assumed annual throughput.

\begin{table*}[t]
\centering
\small
\setlength{\tabcolsep}{4.0pt}
\renewcommand{\arraystretch}{1.14}
\captionof*{table}{\textbf{Table C1. Annual projected API charges and annualized hardware purchase price.} One million questions per year; distinct accounting boundaries, not a like-for-like total-cost-of-ownership comparison.}
\vspace{1pt}
\begin{tabular}{@{}l r r@{}}
\toprule
Workload & Opus API charges / year & Mac Studio purchase price / 3 years \\
\midrule
HotpotQA & \$35,255 & \$3,166.33 \\
2Wiki & \$36,904 & \$3,166.33 \\
MuSiQue & \$44,990 & \$3,166.33 \\
\bottomrule
\end{tabular}
\end{table*}

\begin{figure*}[t]
\centering
\includegraphics[width=\textwidth]{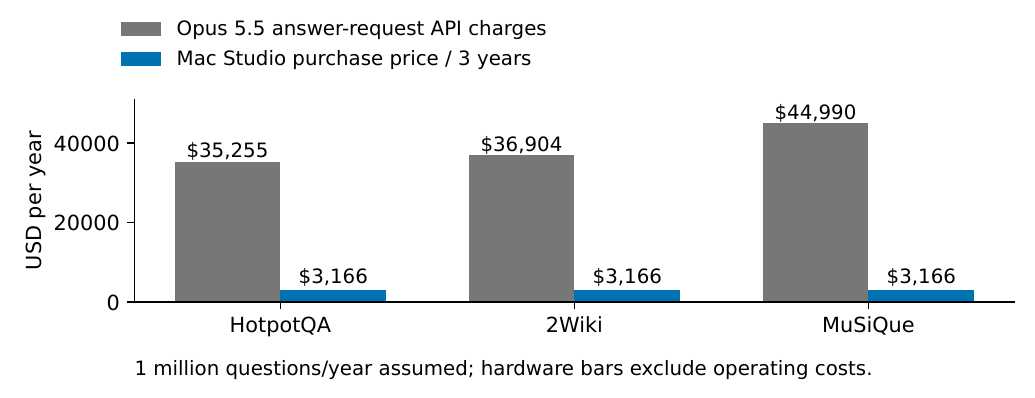}
\caption*{Figure C1. Projected answer-request API charges and annualized hardware acquisition cost. Hardware bars use the cited \$9,499 list price divided by three years. No financing or training-hardware allowance is included. The cloud bars are Opus 5.5 only; Table C3 gives a lower-cost comparator. The comparison does not measure end-to-end cost, local capacity or cost per correct answer.}
\end{figure*}

Cloud charges exclude indexing, retrieval, storage, support, taxes and negotiated, cache or batch discounts. They are projections from recorded usage, not provider invoices. Cheaper cloud configurations exist: Table C3 prices the same workload on Gemini 3.8 Flash, and Section C.3 shows selected-context examples. A complete local-versus-cloud comparison would require comparable workload capacity and all operating costs on both sides.

\subsection*{C.2 Raw cloud usage}
\pdfbookmark[2]{C.2 Raw cloud usage}{c-2-raw-cloud-usage}

\noindent\begin{minipage}{\columnwidth}\centering
\footnotesize
\setlength{\tabcolsep}{4.0pt}
\renewcommand{\arraystretch}{1.14}
\captionof*{table}{\textbf{Table C2. Raw Opus means behind Table C1.} Usage is from 300 locked-test questions per benchmark. Annual figures use unrounded means at the recorded prices.}
\vspace{1pt}
\begin{tabular}{@{}>{\raggedright\arraybackslash}p{39.9pt} >{\centering\arraybackslash}p{46.2pt} >{\centering\arraybackslash}p{48.8pt} >{\centering\arraybackslash}p{74.0pt}@{}}
\toprule
Workload & Mean input tokens & Mean output tokens & API charge / million questions \\
\midrule
HotpotQA & 8,089.18 & 144.93 & \$35,255 \\
2Wiki & 8,456.11 & 153.98 & \$36,904 \\
MuSiQue & 9,677.56 & 313.96 & \$44,990 \\
\bottomrule
\end{tabular}
\end{minipage}\par\vspace{7pt plus 1pt}

\begin{table*}[t]
\centering
\small
\setlength{\tabcolsep}{4.0pt}
\renewcommand{\arraystretch}{1.14}
\captionof*{table}{\textbf{Table C3. The same projection for Gemini 3.8 Flash.} Gemini 3.8 Flash is the highest-scoring cloud comparator on MuSiQue and within one EM point of the highest-scoring cloud comparator on HotpotQA and 2Wiki (Table 2). Usage is from the same 300 locked-test questions per benchmark, at the recorded rates of \$0.50 per million input tokens and \$3 per million output tokens. Billed output tokens include the reasoning tokens reported by the route.}
\vspace{1pt}
\begin{tabular}{@{}l r r r@{}}
\toprule
Workload & Mean input tokens & Mean billed output tokens & API charge / million questions \\
\midrule
HotpotQA & 5,356.73 & 36.48 & \$2,788 \\
2Wiki & 5,704.77 & 88.82 & \$3,119 \\
MuSiQue & 6,424.61 & 206.57 & \$3,832 \\
\bottomrule
\end{tabular}
\end{table*}

At these rates the projected Gemini 3.8 Flash charge for one million questions per year is of the same order as the \$3,166.33 annualized hardware purchase price, and roughly one-twelfth of the Opus projection. The two figures behave differently as volume changes. The API charge is paid per question and scales linearly: at ten million questions per year the same projection is \$27,880 to \$38,320. The hardware is purchased once, and its price does not change with the number of questions; whether a single system sustains a given volume is a throughput question this report does not measure.

Cost is also not the only difference between the two columns. Local hardware keeps documents, queries and model execution on equipment the customer administers, which an external API does not provide at any price, and where policy requires air-gapped operation it is the only deployable option (Section 7.2). The cost comparison therefore depends on which cloud model is the alternative and on the question volume, while the control argument does not.

\subsection*{C.3 Smaller-context reference}
\pdfbookmark[2]{C.3 Smaller-context reference}{c-3-smaller-context-reference}

An October 18 sample used the first 30 development questions per benchmark with selected passages supplied to Opus at the same recorded rates. Its answer-request projections were:

\noindent\begin{minipage}{\columnwidth}\centering
\footnotesize
\setlength{\tabcolsep}{4.0pt}
\renewcommand{\arraystretch}{1.14}
\begin{tabular}{@{}>{\raggedright\arraybackslash}p{44.3pt} >{\centering\arraybackslash}p{57.4pt} >{\centering\arraybackslash}p{48.1pt} >{\centering\arraybackslash}p{54.9pt}@{}}
\toprule
Opus input condition & HotpotQA, USD per million & 2Wiki, USD per million & MuSiQue, USD per million \\
\midrule
STRATUM top five & 7,263 & 9,602 & 11,977 \\
STRATUM up to ten & 12,284 & 13,923 & 16,623 \\
\bottomrule
\end{tabular}
\end{minipage}\par\vspace{7pt plus 1pt}

These rows illustrate the dependence of API cost on context configuration. They use different development questions and are not paired cost/accuracy differences against the locked tests in Section 5.

\subsection*{C.4 Retained artifacts and reproducibility status}
\pdfbookmark[2]{C.4 Retained artifacts and reproducibility status}{c-4-retained-artifacts-and-reproducibility-status}

The headline summaries, paired comparisons and per-run SHA-256 hashes are retained in \texttt{figures\_\allowbreak{}v7/\allowbreak{}STATS\_\allowbreak{}v7.md} and \texttt{figures\_\allowbreak{}v7/\allowbreak{}stats\_\allowbreak{}v7.json}. Figures 2--4 are byte-identical to the v12 graphics built from those frozen statistics. Figure 1 was redrawn to separate the deployment-only candidate-generation path from the evaluated pipeline. Figure C1 uses the unchanged cloud charge projections and the cited hardware list price. Table 5 was recomputed from the same hash-verified locked-test output files; no new inference or bootstrap resampling was performed. In the November 25 revision, the eight-versus-ten paired intervals in Section 4 were computed from the retained development-run outputs with the same bootstrap procedure and seed, which also reproduces the published selection-versus-all-candidates intervals; Table C3 was computed from the retained Gemini 3.8 Flash locked-test outputs; and the MuSiQue sampling description in Section 3.1 was taken from the frozen split manifest and checked by regenerating the 300-question list from its recorded seed. No new inference was run.

The additional cost sample remains \texttt{cost\_\allowbreak{}experiment\_\allowbreak{}2026-10-18.json}. The v13 revision records source hashes, derived counts and verification results alongside its build. The typeset manuscript and figure source package do not constitute a public release of the underlying per-question evaluation records. Independent reproduction has not been established.
\end{document}